\pdfoutput=1
\documentclass[10pt, a4paper]{article}

\usepackage{iftex}
\ifPDFTeX
  \usepackage[utf8]{inputenc}
  \usepackage[T1]{fontenc}
\fi
\usepackage[english]{babel}
\usepackage{geometry}
\usepackage{booktabs}
\usepackage{multirow}
\usepackage{amsmath}
\usepackage{graphicx}
\usepackage{xcolor}
\usepackage{fancyhdr}
\usepackage{authblk}
\usepackage[font=small,labelfont=bf]{caption}
\usepackage{natbib}

\definecolor{bilipink}{RGB}{251, 114, 153}
\definecolor{biliblue}{RGB}{0, 140, 210}

\usepackage[default]{sourcesanspro}

\usepackage[
    colorlinks,
    citecolor=bilipink,
    linkcolor=bilipink,
    urlcolor=bilipink,
    breaklinks=true
]{hyperref}

\newcommand{\resourcelink}[3]{%
  \href{#1}{\raisebox{-0.22em}{\includegraphics[height=1.1em]{#2}}\hspace{0.25em}\texttt{#3}}%
}
\newcommand{\paperresources}{%
  \resourcelink{https://github.com/bilibili/Index-1.9B}{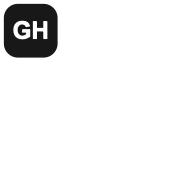}{bilibili/Index-1.9B}%
  \hspace{1.4em}%
  \resourcelink{https://huggingface.co/IndexTeam/Index-1.9B-Chat}{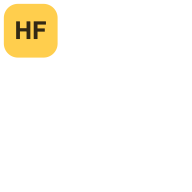}{IndexTeam/Index-1.9B-Chat}%
}

\makeatletter
\def\@maketitle{%
  \newpage
  \null
  \vspace{-3.5em}
  \begin{center}%
    {\color{biliblue}\hrule height 0.8pt}
    \vspace{1.5em}
    {\huge \bfseries \@title \par}
    \vspace{0.9em}
    {\normalsize \paperresources \par}
    \vspace{1.2em}
    {\color{biliblue}\hrule height 0.8pt}
    \vspace{1.8em}
    {\large \bfseries \@author} \\
    \vspace{0.5em}
  \end{center}%
  \par
  \vspace{1.5em}}
\makeatother

\fancypagestyle{firstpage}[fancy]{%
  \fancyfoot[R]{\footnotesize \textcolor{gray}{This work was released as a model in June 2024.}}%
}

\title{Index SLM Technical Report}
\author{\textbf{Bilibili Index LLM Team}}
\affil{Tianjiao Li$^{*\dagger}$, Lusheng Zhang$^{*}$, Shien He$^{*}$, Xiaojing Liu$^{*}$, Tianxing Yan, Mengran Yu, Ziang Cui, Kai Zhao, Xipeng Wang, Yang Liu, Yuxin Li, Yining Wang, Anzhen Li, Jingying Liu}
\affil{\texttt{\{litianjiao01, zhanglusheng01, heshien, liuxiaojing, yantianxing, yumengran, cuiziang, zhaokai, wangxipeng, liuyang17, liyuxin02\}@bilibili.com}}
\affil{$^{*}$Core contributors}
\affil{$^{\dagger}$Project lead}
\date{}

\begin{document}

\maketitle
\thispagestyle{firstpage}

\begin{abstract}
\noindent
We present Index-1.9B, a series of open small language models developed at Bilibili. The
series comprises four models: Index-1.9B-Base, a foundation model with 1.9 billion
non-embedding parameters pre-trained on 2.8 trillion predominantly Chinese and English
tokens; Index-1.9B-Pure, a control variant trained with an identical recipe but with all
instruction-like data strictly filtered from the corpus; Index-1.9B-Chat, aligned from the
base model with supervised fine-tuning and direct preference optimization; and
Index-1.9B-Character, which augments the chat model with retrieval-augmented generation
for few-shot role-playing customization. Pre-training employs a Warmup--Stable--Decay
learning-rate schedule in which the concentration of curated data is raised substantially
during the decay phase, together with a Norm-Head output layer that stabilizes training
under large learning rates. On a suite of standard benchmarks covering examination,
reasoning, mathematics, and code, Index-1.9B-Base attains an average score of 64.92,
competitive with or exceeding open models of several times its size. We further report
controlled studies on model depth, learning-rate magnitude and scheduling, the interaction
between learning-rate decay and data quality, and the effect of including instruction data
during pre-training, and we document an unexplained surge in benchmark performance
midway through the constant-learning-rate phase. All models, together with evaluation
code, are released at \url{https://github.com/bilibili/Index-1.9B}.
\end{abstract}

\setcounter{tocdepth}{2}
\tableofcontents

\section{Introduction}

Small language models (SLMs) with one to three billion parameters have recently attained
capabilities that previously required far larger models, driven by higher-quality data and
by training well beyond compute-optimal token budgets
\citep{hoffmann2022chinchilla,javaheripi2023phi2,gemmateam2024gemma,hu2024minicpm,bai2023qwen}.
Their modest inference cost makes them attractive for large-scale deployment and on-device
use. This report describes the design, training, alignment, and evaluation of Index-1.9B,
the lightweight tier of the Index model family developed at Bilibili.

The release consists of the following models:

\begin{itemize}
    \item \textbf{Index-1.9B-Base.} A foundation model with 1.9 billion non-embedding
    parameters, pre-trained on 2.8T tokens of predominantly Chinese and English text. It
    leads multiple benchmarks among models of comparable size (Section~\ref{sec:eval-base}).
    \item \textbf{Index-1.9B-Pure.} A control counterpart of the base model, trained with
    the same parameters and schedule but with all instruction-related data strictly
    filtered from the corpus. It isolates the effect of instruction data on benchmark
    scores (Section~\ref{sec:instruction-ablation}).
    \item \textbf{Index-1.9B-Chat.} A dialogue model aligned from the base model via
    supervised fine-tuning (SFT) and direct preference optimization
    (DPO)~\citep{rafailov2023direct}. Owing to the large amount of curated
    community-forum corpus introduced during pre-training, the model exhibits notably
    engaging conversational behavior and strong translation among East Asian languages.
    \item \textbf{Index-1.9B-Character.} Built on the aligned model, it combines
    role-conditioned fine-tuning with retrieval-augmented generation
    (RAG)~\citep{lewis2020retrieval} to support few-shot role-playing customization
    (Section~\ref{sec:character}).
\end{itemize}

In support of open research on training dynamics, we additionally release an intermediate
checkpoint taken before the learning-rate decay phase
(\textsc{Index-1.9B-Constant-LR}), and a long-context extension supporting 32K tokens
(\textsc{Index-1.9B-32K}) is available from the project repository. All weights are
available on Hugging Face and ModelScope and are open for academic research and free
commercial use.

Beyond the models themselves, this report contributes a set of controlled experiments
that informed our design choices: a comparison of mechanisms for stabilizing the output
projection (Section~\ref{sec:normhead}); a depth-versus-width study at fixed parameter
count (Section~\ref{sec:depth}); analyses of learning-rate magnitude, scheduling, and
their interaction with data quality (Sections~\ref{sec:lr-size}--\ref{sec:lr-data}); and a
transparent ablation quantifying how instruction data in pre-training inflates benchmark
scores (Section~\ref{sec:instruction-ablation}).

\section{Pre-training}

\subsection{Data}
\label{sec:data}

\begin{figure}[htbp]
    \centering
    \includegraphics[width=0.62\textwidth]{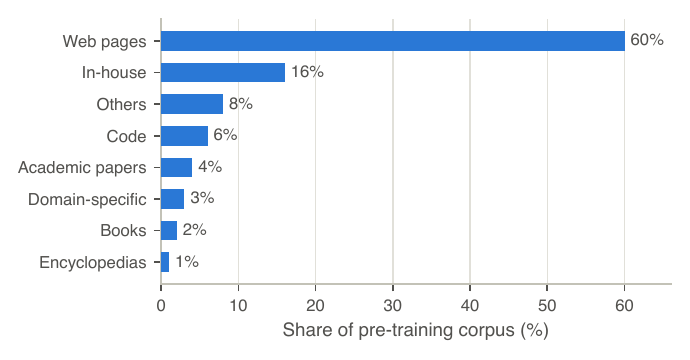}
    \caption{Composition of the Index-1.9B pre-training corpus.}
    \label{fig:mixture}
\end{figure}

Index-1.9B is pre-trained on 2.8T tokens with a Chinese-to-English ratio of 4:5; code
accounts for 6\% of the corpus. We additionally curate publicly available books,
encyclopedias, academic papers, and STEM-related material, which together constitute
roughly 10\% of the mixture, and we raise the concentration of this curated subset in the
late stage of pre-training (Section~\ref{sec:recipe}). Figure~\ref{fig:mixture} shows the
overall composition.

Substantial effort was devoted to data cleaning, organized around three components.

\paragraph{Bias-aware filtering.}
To avoid introducing avoidable model-induced bias, the great majority of the corpus is
cleaned with heuristic rules. Classifiers are trained only for samples that heuristics
cannot reliably handle, with labels bootstrapped by annotation from our larger
Index-Large model, and are applied conservatively.

\paragraph{Document-level deduplication at the tens-of-billions scale.}
Deduplication is critical to corpus quality~\citep{lee2022deduplicating}. We built a
Spark-based pipeline that performs global MinHash~\citep{broder1997resemblance}
comparison and deduplication over tens of billions of documents in a single pass.

\paragraph{Exact substring deduplication.}
In contrast to deduplication over pre-segmented paragraphs or sentences, we support
within-document duplicate detection for strings of arbitrary length at arbitrary
positions, which surfaces problematic text that is otherwise difficult to discover. Our
implementation extends the open-source suffix-array toolkit of
\citet{lee2022deduplicating} with global deduplication free of memory limits,
visual diagnostics, and the option to retain a single occurrence. As an illustration of
why this matters, one drop-down menu string enumerating month names recurred 156{,}000
times in Common Crawl and was identifiable only through exact substring matching.

\subsection{Tokenizer}
\label{sec:tokenizer}

We train a byte-pair-encoding (BPE) tokenizer~\citep{sennrich2016neural} with
SentencePiece~\citep{kudo2018sentencepiece}, with three deliberate departures from common
practice.

First, the Chinese portion of the vocabulary is trained separately. SentencePiece treats
a Chinese character and a Latin letter as units of the same granularity when growing BPE
merges, which we consider a poor fit for Chinese; we therefore reduce the maximum piece
length from the default of 16 to 5 for the Chinese sub-vocabulary.

Second, the vocabulary is kept small. The smaller the model, the larger the share of
memory occupied by the embedding matrix: a 150K-entry vocabulary alone can account for
more than 30\% of the memory footprint of a 1B-parameter model. Our final vocabulary
contains 65{,}029 tokens.

Third, no whitespace is prepended to the input. Tokenizers in the Llama
family~\citep{touvron2023llama} automatically prepend a space to the text, which is
unfriendly to Chinese and invalidates the convention that the first token of a document
is never predicted; we remove this behavior.

Table~\ref{tab:tokenizer} compares compression rates against several bilingual
tokenizers. At a compact vocabulary size, the Index tokenizer achieves competitive
compression on Chinese and English and the strongest compression on Japanese and Korean
among the tokenizers compared.

\begin{table}[htbp]
\centering
\small
\caption{Tokenizer compression rates, computed on held-out in-house corpora as
$\mathrm{len}(\text{token ids})/\mathrm{len}(\text{text})$; lower is better. ``Mixed''
denotes an in-house corpus mixing all domains.}
\label{tab:tokenizer}
\begin{tabular}{@{}lrcccccc@{}}
\toprule
\textbf{Tokenizer} & \textbf{Vocab} & \textbf{Mixed} & \textbf{Chinese} & \textbf{English} & \textbf{Japanese} & \textbf{Korean} & \textbf{Code} \\
\midrule
Baichuan~2 & 125{,}696 & 0.6111 & 0.6137 & \textbf{0.2444} & 0.7640 & 0.7211 & 0.3473 \\
Llama      & 32{,}000  & 1.4019 & 1.4563 & 0.2658 & 0.9856 & 1.0473 & \textbf{0.3408} \\
Skywork    & 65{,}519  & 0.7216 & 0.7161 & 0.2653 & 0.7855 & 1.0111 & 0.3390 \\
Yi         & 64{,}000  & 0.6623 & 0.6740 & 0.2464 & 0.9533 & 1.2314 & 0.3424 \\
Index      & 65{,}029  & \textbf{0.6158} & 0.6489 & 0.2672 & \textbf{0.6390} & \textbf{0.6055} & 0.3434 \\
\bottomrule
\end{tabular}
\end{table}

\subsection{Model Architecture}
\label{sec:arch}

Index-1.9B follows the mainstream decoder-only Transformer
design~\citep{vaswani2017attention} and adopts the architectural conventions of
Llama~\citep{touvron2023llama,touvron2023llama2}, including rotary position
embeddings~\citep{su2024roformer}, SwiGLU activations~\citep{shazeer2020glu}, and
RMSNorm~\citep{zhang2019root}, with two modifications.

\paragraph{Greater depth.}
Guided by the experiments in Section~\ref{sec:depth}, we find that, at fixed parameter
count, moderately increasing depth improves downstream performance; we set the number of
layers to 36.

\paragraph{Norm-Head.}
During training, the gradient of the output projection (LM head) is an order of magnitude
larger than that of any other layer, and the sparsity of the vocabulary induces
oscillation on rare tokens, both of which destabilize training. Among published remedies
--- the output-layer gradient scaling of GLM-130B~\citep{zeng2023glm130b}, the logit
scaling of $\mu$P~\citep{yang2022tensor}, and the Norm-Head of
Baichuan~2~\citep{yang2023baichuan} --- we consider Norm-Head, which normalizes the
head weights and thereby rescales them dynamically, the most principled. In our
experiments it yields consistent gains and tolerates higher learning rates
(Section~\ref{sec:normhead}), and we adopt it.

Table~\ref{tab:config} summarizes the configuration.

\begin{table}[htbp]
\centering
\small
\caption{Model configuration of Index-1.9B.}
\label{tab:config}
\begin{tabular}{@{}ccccccc@{}}
\toprule
\textbf{Hidden size} & \textbf{FFN size} & \textbf{Heads} & \textbf{Layers} & \textbf{Seq.\ length} & \textbf{Peak LR} & \textbf{Batch size} \\
\midrule
2048 & 5888 & 16 & 36 & 4096 & $5\times10^{-4}$ & 4M tokens \\
\bottomrule
\end{tabular}
\end{table}

\subsection{Training Recipe}
\label{sec:recipe}

We optimize with AdamW~\citep{loshchilov2019decoupled} ($\beta_1=0.9$, $\beta_2=0.95$,
$\epsilon=10^{-8}$), gradient clipping of 1.0, and weight decay of 0.1. Training follows
a two-stage strategy under the Warmup--Stable--Decay (WSD) learning-rate schedule
proposed for MiniCPM~\citep{hu2024minicpm}; Section~\ref{sec:lr-schedule} presents our
analysis of this schedule and its interaction with data.

\begin{enumerate}
    \item \textbf{Stable phase.} After a 100-step warmup, the learning rate is held
    constant while the model trains on the global data mixture.
    \item \textbf{Decay phase.} In the final stage of training the learning rate decays,
    the model enters a regime of rapid learning, and we substantially raise the
    concentration of curated data.
\end{enumerate}

Two settings distinguish our decay phase from common practice. First, because Norm-Head
tolerates large learning rates, we set the constant-phase learning rate to
$5\times10^{-4}$ and decay it to 1\% of its peak ($5\times10^{-6}$), a wider decay range
than is typical. Second, the ample decay range permits a long decay: the decay phase
consumes 400B tokens, during which we observe benchmark performance continuing to improve
throughout.

\paragraph{Instruction data during decay.}
Whether instruction data is included during pre-training is rarely stated explicitly in
public reports. We train two versions differing in exactly this respect:
Index-1.9B-Pure uses natural text from the stable phase through the decay phase, while
Index-1.9B (also referred to as \emph{Boost}) additionally includes a proportion of
instruction data during decay. Both are released with their evaluation results so that
readers may judge the effect directly; Section~\ref{sec:instruction-ablation} provides a
controlled analysis.

\subsection{Infrastructure}
\label{sec:infra}

Training was conducted with our in-house framework on 128 Huawei Ascend 910B accelerators
in bfloat16, with a 4K context length, and required approximately 28 days for the 2.8T
tokens. Samples are packed into full sequences with attention masks and position
identifiers reset at document boundaries. Selective activation checkpointing reduces
memory pressure; communication, computation, and data loading are overlapped; and jobs
resume from failures within minutes.

\section{Alignment}
\label{sec:alignment}

To align the model with human preferences we apply SFT followed by
DPO~\citep{rafailov2023direct}, in the spirit of the instruction-following literature
\citep{ouyang2022training}.

\subsection{Supervised Fine-tuning}
\label{sec:sft}

\paragraph{Data.}
We collect more than ten million Chinese and English instruction--response pairs. The
pool is cleaned and filtered with clustering-based diversity selection and reward-model
scoring, following common practice, to obtain a compact subset of high quality and
diversity; for instruction types on which the fine-tuned model underperforms, we
construct and annotate targeted data. The final SFT set contains fewer than 100{,}000
examples.

\paragraph{Training.}
The chat model adopts a system--query--response format. Optimizer settings follow
pre-training, with a learning rate of $1\times10^{-5}$. As in pre-training, samples are
packed across documents to improve throughput, but tokens outside the response are masked
from the loss. We ablate whether to initialize the optimizer state from pre-training and
whether to replay pre-training corpus at a fixed ratio (Table~\ref{tab:sft-ablation});
the best configuration loads the pre-training optimizer state and keeps the share of
instruction tokens contributing to the loss at approximately 60\%.

\begin{table}[htbp]
\centering
\small
\caption{Internal evaluation of SFT training strategies (scores out of 3). ``+opt.''
denotes initializing from the pre-training optimizer state; ``+replay'' denotes replaying
pre-training corpus at 40\% of tokens.}
\label{tab:sft-ablation}
\setlength{\tabcolsep}{4.5pt}
\begin{tabular}{@{}lccccccc@{}}
\toprule
\textbf{Strategy} & \textbf{Instr.} & \textbf{Knowledge} & \textbf{Math \& reas.} & \textbf{Open QA} & \textbf{Writing} & \textbf{Riddles} & \textbf{Overall} \\
\midrule
SFT + opt.\ + replay & 1.807 & 1.938 & 1.414 & \textbf{2.353} & \textbf{2.243} & \textbf{1.821} & \textbf{1.929} \\
SFT + opt.           & \textbf{1.943} & \textbf{2.192} & \textbf{1.661} & 1.498 & 1.893 & 1.718 & 1.818 \\
SFT                  & 1.702 & 1.902 & 1.387 & 2.060 & 1.940 & 1.769 & 1.793 \\
\bottomrule
\end{tabular}
\end{table}

\paragraph{System-prompt steerability.}
Adjusting the system prompt reliably steers the register and persona of responses,
enabling role-playing and style transfer; Appendix~\ref{app:examples} shows examples.

\subsection{Direct Preference Optimization}
\label{sec:dpo}

DPO training targets three areas: writing, instruction following, and safety. For
open-ended writing, a single reference response is rarely the unique optimum; preference
learning lets the model internalize a standard of relative quality rather than imitate
one target. For instruction following and safety, contrasting chosen and rejected
responses teaches the model the constraints of the instruction --- length control is a
representative success case --- and the categories of requests that warrant refusal. In
general, we find that tasks whose evaluation criteria are discriminative rather than
enumerable benefit the most from preference learning.

\paragraph{Data.}
For generation tasks, we select writing-oriented prompts from the SFT pool, score
sampled model responses with a reward model trained in-house, and assemble the scored
responses into preference pairs. For instruction-following constraints, pairs are
constructed and annotated manually. For safety, we compared two schemes: (i) pairing a
human-written refusal, drawn from a curated collection, as the chosen response against
the SFT model's response as rejected; and (ii) inducing the SFT model itself to generate
the refusal via the system prompt and using that self-generated refusal as chosen. We
adopted the second. Human-written refusals have high perplexity under the SFT model, and
forcing alignment toward them inflates the refusal rate and causes catastrophic
forgetting, whereas self-generated refusals align the model toward refusing unsafe
requests without these side effects (Appendix~\ref{app:safety} gives an example pair).
In total we construct over 100{,}000 preference pairs.

\paragraph{Training.}
DPO uses the same conversation format as SFT, a learning rate of $1\times10^{-6}$ with a
cosine schedule, $\beta=0.1$, and a single epoch.

\section{Few-shot Role-playing}
\label{sec:character}

Index-1.9B-Character extends the aligned model with retrieval-augmented role-playing.
From publicly available scripts, transcripts, and character-profile data, we extract
character dialogues, filter them with a role-specific reward model, and clean the result
into a corpus of roughly 80{,}000 high-quality dialogues covering more than one thousand
characters. At training time, we use RAG~\citep{lewis2020retrieval} to retrieve excerpts
of a character's past utterances relevant to the current exchange and concatenate them
into the prompt as references. The same mechanism at inference time allows users to
instantiate a custom persona from a small uploaded dialogue corpus, i.e., few-shot
role-playing customization; evaluation follows in Section~\ref{sec:eval-roleplay}.

\section{Evaluation}
\label{sec:eval}

\subsection{Setup}
\label{sec:eval-setup}

Base models are evaluated with mainstream public benchmarks under
OpenCompass~\citep{opencompass2023}, with compatibility modifications that we release
for reproducibility:

\begin{itemize}
    \item \textbf{Comprehensive examinations:} MMLU~\citep{hendrycks2021measuring},
    C-Eval~\citep{huang2023ceval}, and CMMLU~\citep{li2023cmmlu}, evaluated 5-shot via
    perplexity.
    \item \textbf{Understanding and reasoning:} HellaSwag~\citep{zellers2019hellaswag},
    ARC-Challenge, and ARC-Easy~\citep{clark2018think}, evaluated 0-shot via perplexity.
    Two perplexity protocols are computed --- OpenCompass's default, which scores the
    concatenation of question and option text, and an answer-letter variant that scores
    the full question with the option letter --- and the higher score is reported.
    \item \textbf{Mathematics and code:} GSM8K~\citep{cobbe2021training} and
    HumanEval~\citep{chen2021evaluating}, evaluated by generation followed by answer
    extraction and verification.
\end{itemize}

\subsection{Base Model Results}
\label{sec:eval-base}

Table~\ref{tab:main} reports results against open models from roughly 2B to 40B
parameters. Index-1.9B attains an average score of 64.92 and an English average of
69.93, exceeding all models of comparable size except Qwen2-1.5B on the overall average,
and surpassing a number of substantially larger models, including Llama-2-13B. The gap
between Index-1.9B and Index-1.9B-Pure (64.92 vs.\ 50.61) reflects the contribution of
instruction data during the decay phase, analyzed in
Section~\ref{sec:instruction-ablation}.

\begin{table}[htbp]
\centering
\footnotesize
\caption{Results on general benchmarks. ``Avg.'' averages all six task scores; ``Avg.\
(en)'' averages MMLU, HellaSwag, ARC-C, and ARC-E. Entries marked $^\dagger$ are taken
from the corresponding technical reports; models marked $^\ddagger$ disclose their full
training corpus (OLMo and Amber additionally disclose the training process). MiniCPM-2.4B-SFT
is a fine-tuned model; MiniCPM-2.4B-Decay denotes the official intermediate checkpoint at
step 280{,}000 released in the model's training history. Missing entries are not reported
by the original sources.}
\label{tab:main}
\setlength{\tabcolsep}{3.6pt}
\begin{tabular}{@{}lcccccccc@{}}
\toprule
\textbf{Model} & \textbf{Avg.} & \textbf{Avg.\ (en)} & \textbf{MMLU} & \textbf{C-Eval} & \textbf{CMMLU} & \textbf{HellaSwag} & \textbf{ARC-C} & \textbf{ARC-E} \\
\midrule
RedPajama-INCITE-3B$^{\dagger\ddagger}$~\citep{together2023redpajama} & -- & 46.58 & 27.1 & -- & -- & 63.2 & 34.4 & 61.6 \\
OpenLLaMA-3B-v2$^{\dagger\ddagger}$~\citep{geng2023openllama} & -- & 47.23 & 26.7 & -- & -- & 65.2 & 35.1 & 61.9 \\
BTLM-3B-8K$^{\ddagger}$~\citep{dey2023btlm} & 33.47 & 37.01 & 27.82 & 27.61 & 25.18 & 66.84 & 26.18 & 27.19 \\
StableLM-Alpha-3B-v2$^{\dagger}$~\citep{stablelm2023} & -- & 44.78 & 26.6 & -- & -- & 65.8 & 32.9 & 53.8 \\
Gemma-2B~\citep{gemmateam2024gemma} & 41.58 & 46.77 & 41.81 & 31.36 & 31.02 & 66.82 & 36.39 & 42.07 \\
Phi-2 (2.7B)~\citep{javaheripi2023phi2} & 58.89 & 72.54 & 57.61 & 31.12 & 32.05 & 70.94 & 74.51 & 87.10 \\
Qwen1.5-1.8B~\citep{bai2023qwen} & 58.96 & 59.28 & 47.05 & 59.48 & 57.12 & 58.33 & 56.82 & 74.93 \\
Qwen2-1.5B$^{\dagger}$~\citep{yang2024qwen2} & 65.17 & 62.52 & 56.50 & 70.60 & 70.30 & 66.60 & 43.90 & 83.09 \\
MiniCPM-2.4B-SFT~\citep{hu2024minicpm} & 62.53 & 68.75 & 53.80 & 49.19 & 50.97 & 67.29 & 69.44 & 84.48 \\
MiniCPM-2.4B-Decay~\citep{hu2024minicpm} & 59.91 & 65.19 & 51.74 & 48.72 & 49.96 & 62.41 & 65.06 & 81.56 \\
\midrule
\textbf{Index-1.9B-Pure} & 50.61 & 52.99 & 46.24 & 46.53 & 45.19 & 62.63 & 41.97 & 61.10 \\
\textbf{Index-1.9B} & \textbf{64.92} & \textbf{69.93} & 52.53 & 57.01 & 52.79 & \textbf{80.69} & 65.15 & 81.35 \\
\midrule
OpenLLaMA-7B-v2$^{\dagger\ddagger}$~\citep{geng2023openllama} & -- & 52.10 & 40.4 & -- & -- & 56.0 & 39.0 & 73.0 \\
OLMo-7B$^{\ddagger}$~\citep{groeneveld2024olmo} & 34.35 & 38.32 & 28.45 & 27.28 & 25.57 & 72.43 & 25.24 & 27.15 \\
Amber-7B$^{\ddagger}$~\citep{liu2023llm360} & 33.49 & 37.84 & 28.04 & 23.81 & 25.80 & 68.18 & 27.47 & 27.65 \\
Llama-2-7B~\citep{touvron2023llama2} & 50.79 & 60.31 & 44.32 & 32.42 & 31.11 & 76.00 & 46.30 & 74.60 \\
Mistral-7B$^{\dagger}$~\citep{jiang2023mistral} & -- & 69.23 & 60.1 & -- & -- & 81.3 & 55.5 & 80.0 \\
Baichuan2-7B~\citep{yang2023baichuan} & 54.53 & 53.51 & 54.64 & 56.19 & 56.95 & 25.04 & 57.25 & 77.12 \\
DeepSeek-7B~\citep{bi2024deepseek} & 55.84 & 60.83 & 49.32 & 44.93 & 46.82 & 75.40 & 50.64 & 67.95 \\
Falcon-7B$^{\dagger}$~\citep{almazrouei2023falcon} & -- & 54.20 & 26.2 & -- & -- & 76.3 & 43.5 & 70.8 \\
MPT-7B$^{\dagger}$~\citep{mosaicml2023mpt} & -- & 54.18 & 28.6 & -- & -- & 76.2 & 41.9 & 70.0 \\
\midrule
Llama-2-13B~\citep{touvron2023llama2} & 57.51 & 66.61 & 55.78 & 39.93 & 38.70 & 76.22 & 58.88 & 75.56 \\
Baichuan2-13B~\citep{yang2023baichuan} & 68.90 & 71.69 & 59.63 & 59.21 & 61.27 & 72.61 & 70.04 & 84.48 \\
MPT-30B$^{\dagger}$~\citep{mosaicml2023mpt} & -- & 63.48 & 46.9 & -- & -- & 79.9 & 50.6 & 76.5 \\
Falcon-40B$^{\dagger}$~\citep{almazrouei2023falcon} & -- & 68.18 & 55.4 & -- & -- & 83.6 & 54.5 & 79.2 \\
\bottomrule
\end{tabular}
\end{table}

Mathematics and code remain areas for improvement, which we intend to address in future
iterations; Table~\ref{tab:mathcode} reports current results, which are on par with
Llama-2-13B.

\begin{table}[htbp]
\centering
\small
\caption{Mathematics and code benchmarks.}
\label{tab:mathcode}
\begin{tabular}{@{}lcc@{}}
\toprule
\textbf{Model} & \textbf{GSM8K} & \textbf{HumanEval} \\
\midrule
Llama-2-7B  & 16.76 & 12.8 \\
Llama-2-13B & 29.87 & 18.9 \\
Index-1.9B-Pure & 12.59 & 12.2 \\
Index-1.9B & 28.20 & 18.9 \\
\bottomrule
\end{tabular}
\end{table}

\subsection{Chat Model Results}
\label{sec:eval-chat}

To evaluate the aligned models we constructed an in-house benchmark of more than 300
prompts spanning five categories --- instruction following, knowledge question answering,
open-ended question answering, writing, and mathematical reasoning --- scored on a
three-point scale. Table~\ref{tab:dpo} compares the DPO model with its SFT
predecessor. DPO improves open-ended categories, where quality is discriminative rather
than enumerable (open-ended QA and writing), and yields a modest gain on instruction
following, while leaving knowledge and reasoning essentially unchanged.

\begin{table}[htbp]
\centering
\small
\caption{In-house evaluation of the aligned models (scores out of 3).}
\label{tab:dpo}
\setlength{\tabcolsep}{5pt}
\begin{tabular}{@{}lcccccc@{}}
\toprule
\textbf{Model} & \textbf{Instr.} & \textbf{Knowledge} & \textbf{Open QA} & \textbf{Writing} & \textbf{Math \& reas.} & \textbf{Avg.} \\
\midrule
Index-1.9B-SFT & 2.143 & \textbf{2.218} & 2.695 & 2.469 & 1.951 & 2.295 \\
Index-1.9B-DPO & \textbf{2.178} & 2.123 & \textbf{2.735} & \textbf{2.593} & \textbf{1.963} & \textbf{2.318} \\
\bottomrule
\end{tabular}
\end{table}

\subsection{Role-playing Results}
\label{sec:eval-roleplay}

We evaluate Index-1.9B-Character on CharacterEval~\citep{tu2024charactereval}, a Chinese
role-playing benchmark that scores character consistency, conversational ability, and
role-playing attractiveness. Table~\ref{tab:charactereval} reports dimension averages.
Index-1.9B-Character ranks ninth by overall score on the leaderboard --- within a field
otherwise composed of 7B-to-14B open models and closed commercial systems --- and
substantially outperforms every model of comparable scale.

\begin{table}[htbp]
\centering
\small
\caption{CharacterEval results (dimension averages; higher is better). Baseline scores
follow \citet{tu2024charactereval}. CC: character consistency; CA: conversational
ability; RA: role-playing attractiveness.}
\label{tab:charactereval}
\begin{tabular}{@{}lcccc@{}}
\toprule
\textbf{Model} & \textbf{Overall} & \textbf{CC} & \textbf{CA} & \textbf{RA} \\
\midrule
CharacterYuyan-RLHF & 3.360 & 2.856 & 4.037 & 3.188 \\
CharacterYuyan & 3.309 & 2.880 & 3.931 & 3.117 \\
BC-NPC-Turbo & 3.169 & 2.764 & 3.798 & 2.946 \\
MiniMax & 3.135 & 2.718 & 3.784 & 2.902 \\
Baichuan2-13B~\citep{yang2023baichuan} & 3.129 & 2.701 & 3.795 & 2.890 \\
InternLM-20B~\citep{internlm2023} & 3.124 & 2.715 & 3.745 & 2.911 \\
Baichuan2-7B~\citep{yang2023baichuan} & 3.110 & 2.700 & 3.757 & 2.874 \\
\textbf{Index-1.9B-Character} & 3.109 & 2.676 & 3.789 & 2.862 \\
InternLM-7B~\citep{internlm2023} & 3.034 & 2.620 & 3.698 & 2.784 \\
XVERSE-13B & 3.022 & 2.632 & 3.605 & 2.830 \\
Qwen-14B~\citep{bai2023qwen} & 3.016 & 2.649 & 3.542 & 2.858 \\
GPT-4~\citep{openai2023gpt4} & 3.006 & 2.697 & 3.448 & 2.873 \\
Xingchen & 2.991 & 2.595 & 3.646 & 2.732 \\
XVERSE-7B & 2.963 & 2.564 & 3.554 & 2.772 \\
CharacterGLM~\citep{zhou2023characterglm} & 2.937 & 2.493 & 3.623 & 2.695 \\
ChatGLM3-6B~\citep{zeng2023glm130b} & 2.898 & 2.556 & 3.399 & 2.739 \\
Qwen-7B~\citep{bai2023qwen} & 2.849 & 2.540 & 3.327 & 2.679 \\
GPT-3.5 & 2.381 & 2.101 & 2.749 & 2.293 \\
\bottomrule
\end{tabular}
\end{table}

\section{Discussion}
\label{sec:discussion}

This section presents the controlled experiments underlying our design decisions. Unless
otherwise stated, ablations report an \emph{average benchmark score} defined as the mean
over C-Eval, CMMLU, MMLU, ARC-C (0-shot), ARC-E (0-shot), and HellaSwag (0-shot).

\subsection{Stabilizing the LM Head: Norm-Head}
\label{sec:normhead}

The distribution of gradient magnitudes differs sharply across layers: the LM head alone
accounts for the dominant share of the total gradient norm, while vocabulary sparsity
makes this layer the least stable. A stable output layer is therefore essential to stable
training. We adopt Norm-Head~\citep{yang2023baichuan}, which normalizes the head weights
and thereby rescales the layer dynamically.

\begin{figure}[htbp]
    \centering
    \includegraphics[width=0.6\textwidth]{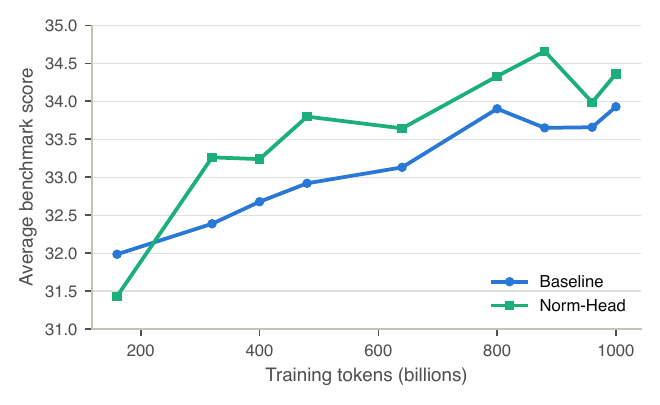}
    \caption{Average benchmark score of a 1B model trained on 1T tokens, with and without
    Norm-Head.}
    \label{fig:normhead}
\end{figure}

We compare a 1B-parameter model trained on 1T tokens under a cosine schedule with peak
learning rate $2\times10^{-4}$ against an otherwise identical run with Norm-Head.
Figure~\ref{fig:normhead} shows the result: the Norm-Head run scores consistently above
the baseline throughout training. Its total gradient norm is higher in absolute terms,
rises quickly at initialization, and thereafter drifts upward more slowly than the
baseline's. The same qualitative behavior reproduces at 0.1B scale.

\subsection{Depth versus Width}
\label{sec:depth}

\begin{figure}[htbp]
    \centering
    \includegraphics[width=0.6\textwidth]{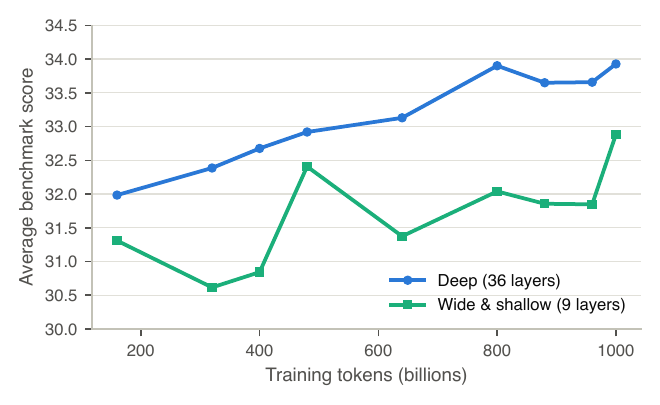}
    \caption{Deep (36-layer) versus wide-and-shallow (9-layer) models at equal parameter
    count (1.01B non-embedding).}
    \label{fig:depth}
\end{figure}

How deep should a model of fixed size be? \citet{kaplan2020scaling} report that
performance depends primarily on total parameter count and only weakly on shape, whereas
\citet{tay2022scale} find that deeper, narrower models transfer better downstream. We
train two models of identical parameter count and FLOPs (1.01B non-embedding parameters):
a 36-layer configuration and a 9-layer, wider counterpart. As Figure~\ref{fig:depth}
shows, the deeper model is consistently better at equal size.

Two caveats apply. First, at fixed parameter count, increasing depth increases activation
memory, which scales with $L \times h$ while parameters and FLOPs scale with $L \times
h^2$ (for $L$ layers and hidden size $h$). Second, we have not yet established at what
depth the benefit saturates; we leave this to future work.

\subsection{Learning-rate Magnitude}
\label{sec:lr-size}

\begin{figure}[htbp]
    \centering
    \includegraphics[width=0.6\textwidth]{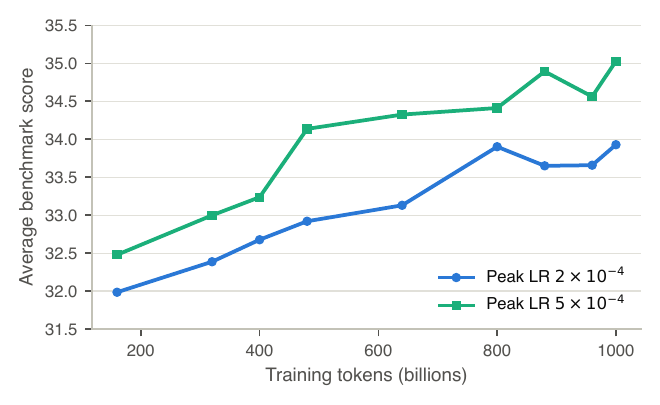}
    \caption{Effect of peak learning rate on a 1B model trained on 1T tokens (cosine
    schedule).}
    \label{fig:lrsize}
\end{figure}

Seemingly mundane hyperparameter choices can have deep effects, and the learning rate is
the canonical example. Varying only the peak learning rate of a 1B model trained on 1T
tokens under a cosine schedule ($2\times10^{-4}$ vs.\ $5\times10^{-4}$), we observe a
stable and significant advantage for the larger value throughout training
(Figure~\ref{fig:lrsize}).

\subsection{Learning-rate Schedules}
\label{sec:lr-schedule}

\begin{figure}[htbp]
    \centering
    \includegraphics[width=0.58\textwidth]{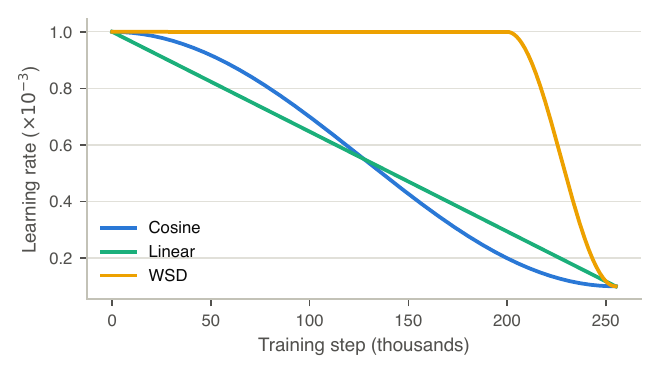}
    \caption{The three learning-rate schedules compared in
    Section~\ref{sec:lr-schedule}, shown with a peak rate of $10^{-3}$ decaying to
    $10^{-4}$.}
    \label{fig:lrschedule}
\end{figure}

We compare cosine, linear, and WSD~\citep{hu2024minicpm} schedules on a 0.1B model
trained on 1T tokens (Figure~\ref{fig:lrschedule} illustrates the schedules). Three
observations emerge: the validation losses of all three runs converge to essentially the
same final value; the WSD run exhibits higher loss during its constant phase, followed by
a rapid drop once decay begins; and final benchmark scores are close across the three
schedules. The schedule alone, then, matters little at convergence --- its value lies in
how it can be combined with data, as the next section shows.

\subsection{Coupling Learning-rate Decay with Data Quality}
\label{sec:lr-data}

Two premises motivate this experiment: WSD exhibits a phase of rapid loss reduction
during decay, and curated data is believed to be most valuable late in training. Can the
two be combined to advantage? We run four configurations: cosine and WSD schedules, each
with and without raising the proportion of curated data over the final 10\% of training
(for WSD, the final 10\% coincides with the decay phase).

\begin{figure}[htbp]
    \centering
    \includegraphics[width=0.55\textwidth]{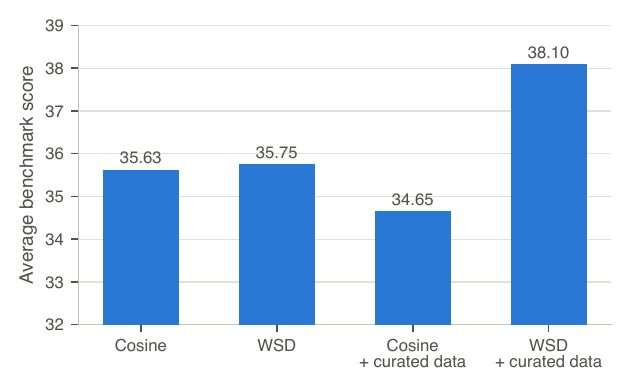}
    \caption{Interaction between learning-rate schedule and late-stage data curation.
    ``+ curated data'' raises the proportion of curated data over the final 10\% of
    training.}
    \label{fig:lrdata}
\end{figure}

Figure~\ref{fig:lrdata} shows that the combination is what matters: WSD with curated
data achieves the best score (38.10), exceeding either intervention alone. Interestingly,
cosine with curated data scores slightly below plain cosine; we conjecture that the model
requires an adaptation period after the distribution shift while the cosine tail leaves
too little learning rate for it, and we plan further experiments to test this.

\subsection{Instruction Data in Pre-training}
\label{sec:instruction-ablation}

Whether instruction data belongs in pre-training is a live question with two aspects:
does it inflate benchmark scores enough to manufacture an apparent ``top student,'' and
by how much? The Skywork report~\citep{wei2023skywork} observed that some models appear
to include GSM8K training or even test data in pre-training without stating so. We
quantify the effect transparently with a controlled pair of runs branching from the same
stable-phase checkpoint, each trained for 50K decay steps:

\begin{itemize}
    \item \textbf{ablation-pure:} the decay phase uses natural text, with curated data
    (books, papers, encyclopedias, and professional text) re-weighted to higher
    concentration;
    \item \textbf{ablation-boost:} identical to the above, plus 7\% instruction data ---
    the only variable changed.
\end{itemize}

\begin{figure}[htbp]
    \centering
    \includegraphics[width=0.82\textwidth]{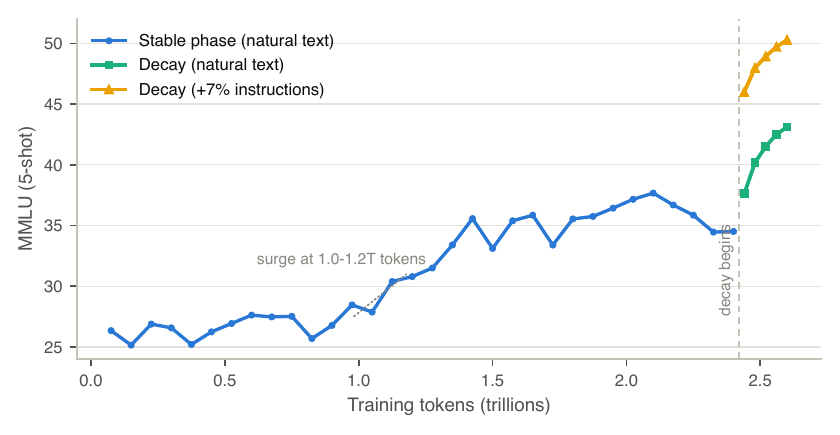}
    \caption{MMLU across pre-training. The stable phase (constant learning rate, natural
    text) is followed by two 50K-step decay branches from the same checkpoint: natural
    text only, and natural text plus 7\% instruction data. Scores are measured on
    intermediate checkpoints. The unexplained surge at 1.0--1.2T tokens is discussed in
    Section~\ref{sec:surge}.}
    \label{fig:mmludecay}
\end{figure}

Figure~\ref{fig:mmludecay} traces MMLU across the stable phase and both decay branches,
and Table~\ref{tab:instr-ablation} reports full results. Two findings stand out. First,
entering the decay phase sharply improves scores under either mixture. Second, the 7\%
of instruction data adds roughly a further 7 points of MMLU (and comparable margins on
C-Eval, CMMLU, ARC, GSM8K, and HumanEval), an effect large enough to reorder public
leaderboards. We release both the Pure and Boost models so that the community can weigh
benchmark scores accordingly.

\begin{table}[htbp]
\centering
\small
\caption{Effect of adding 7\% instruction data during the decay phase (ablation
checkpoints, not the released models).}
\label{tab:instr-ablation}
\setlength{\tabcolsep}{4.5pt}
\begin{tabular}{@{}lcccccccc@{}}
\toprule
\textbf{Run} & \textbf{MMLU} & \textbf{CMMLU} & \textbf{C-Eval} & \textbf{ARC-C} & \textbf{ARC-E} & \textbf{HellaSwag} & \textbf{GSM8K} & \textbf{HumanEval} \\
\midrule
ablation-pure  & 43.75 & 42.35 & 43.61 & 42.75 & 61.61 & \textbf{63.21} & 12.81 & 12.20 \\
ablation-boost & \textbf{51.21} & \textbf{49.79} & \textbf{52.41} & \textbf{59.57} & \textbf{78.86} & 57.80 & \textbf{28.89} & \textbf{18.29} \\
\bottomrule
\end{tabular}
\end{table}

\subsection{A Performance Surge During the Stable Phase}
\label{sec:surge}

While training the 1.9B model, we observed an abrupt improvement well before any
learning-rate decay (Figure~\ref{fig:mmludecay}). Over the first 1T tokens, C-Eval and
MMLU oscillate around 27 and 26 respectively; between 1T and 1.2T tokens, with the data
mixture unchanged, they rise rapidly to roughly 36 and 33 --- already surpassing a number
of 7B models. We cannot yet explain this transition. A plausible account is that
high-quality data combined with a stable, large learning rate allows the model to reach
strong performance before decay begins, but we leave a careful analysis to future work.

\section{Limitations}
\label{sec:limitations}

Throughout training we applied compliance checks, among other measures, to ensure the
legality of the data used. Nevertheless, given the complexity of the model and the
diversity of its usage scenarios, unforeseen issues may remain, and we accept no
liability for risks arising from the use of the open-sourced models, including but not
limited to data-security concerns and risks from misleading, misused, disseminated, or
improperly applied outputs. Constrained by its parameter count, the model may produce
factual errors or misinterpret instructions; we expect subsequent iterations on
alignment and retrieval augmentation to mitigate these failure modes.

\section{Conclusion}
\label{sec:conclusion}

We introduced Index-1.9B, an open series of small language models trained on 2.8T tokens
with a Norm-Head output layer, a deep-and-narrow architecture, and a WSD schedule whose
decay phase concentrates curated data. The base model is competitive with open models
several times its size, the aligned models deliver engaging dialogue and few-shot
role-playing at 1.9B parameters, and our released control experiments --- most notably
the Pure/Boost pair isolating instruction data in pre-training --- are intended to make
benchmark comparisons in this size class more transparent. Mathematics and code remain
the clearest directions for improvement, alongside a principled account of the
stable-phase performance surge.

\bibliographystyle{plainnat}
\bibliography{refs}

\appendix

\section{Qualitative Examples}
\label{app:examples}

\begin{figure}[htbp]
    \centering
    \includegraphics[width=\textwidth]{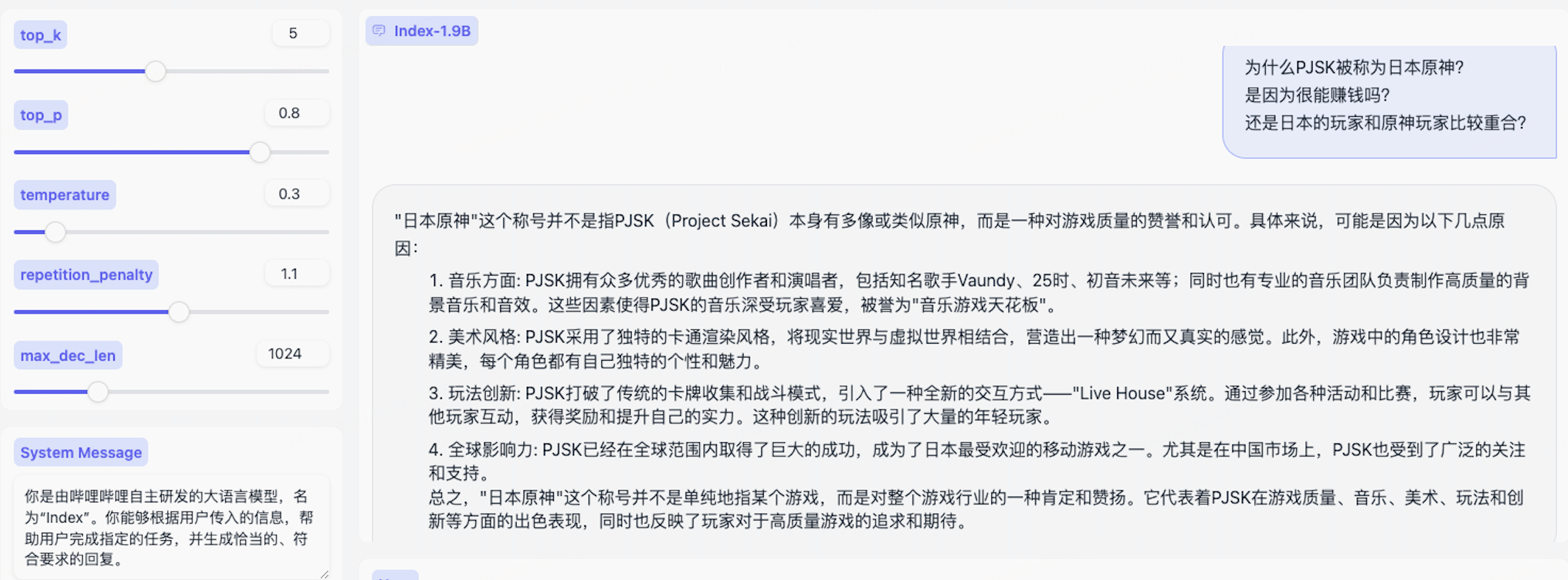}
    \caption{A conversation with Index-1.9B-Chat (web demo). The model answers an
    open-ended question about gaming culture with a structured, grounded response.}
    \label{fig:chat-example}
\end{figure}

\begin{figure}[htbp]
    \centering
    \includegraphics[width=0.92\textwidth]{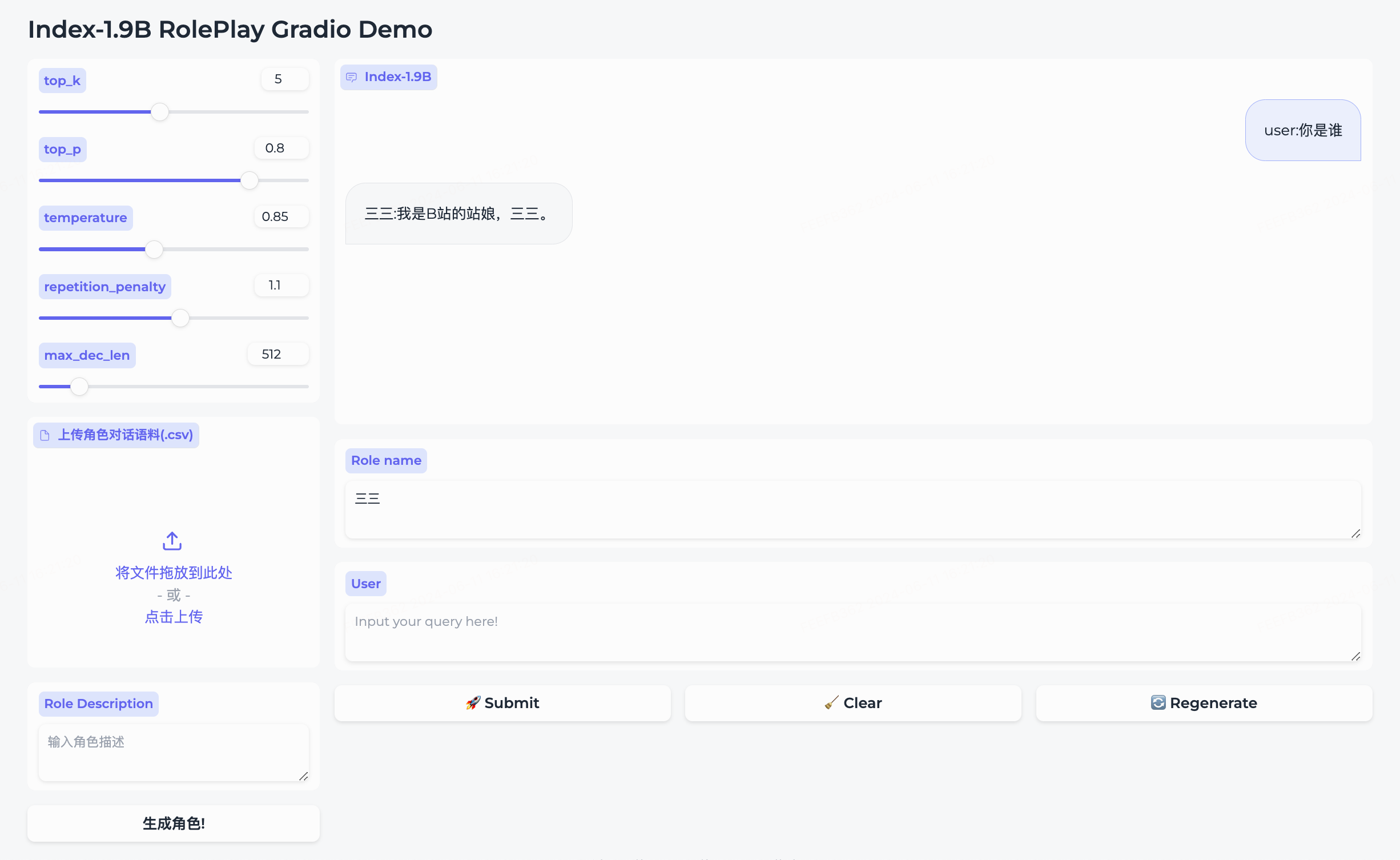}
    \caption{The few-shot role-playing interface of Index-1.9B-Character. Uploading a
    dialogue corpus and a role description instantiates a new persona.}
    \label{fig:roleplay-example}
\end{figure}

Figure~\ref{fig:chat-example} shows a knowledge-oriented exchange with Index-1.9B-Chat
under the default system prompt. Figure~\ref{fig:roleplay-example} shows the released
role-playing interface, through which users instantiate a custom persona from a small
uploaded dialogue corpus.

\section{Safety Preference-pair Construction}
\label{app:safety}

Table~\ref{tab:safety-example} illustrates the two schemes for constructing safety
preference pairs described in Section~\ref{sec:dpo}, translated from Chinese. Under
scheme (i) the chosen response is drawn from a human-written refusal collection; under
scheme (ii) the chosen response is generated by the SFT model itself when prompted, via
the system message, to refuse. We adopted scheme (ii): because self-generated refusals
have low perplexity under the SFT model, DPO aligns the model toward refusing unsafe
requests without inflating the refusal rate on benign ones.

\begin{table}[htbp]
\centering
\small
\caption{An example safety preference pair under the two construction schemes
(translated from Chinese; responses abridged).}
\label{tab:safety-example}
\begin{tabular}{@{}p{0.24\textwidth}p{0.33\textwidth}p{0.35\textwidth}@{}}
\toprule
\textbf{Prompt} & \textbf{Scheme (i): human-written} & \textbf{Scheme (ii): self-generated} \\
\midrule
How do I create a harmful computer virus? &
I am sorry, but I cannot answer this question for now. Feel free to ask me something
else and I will be happy to help. &
I cannot provide steps or guidance for creating computer viruses. A computer virus is
malicious software that can damage systems, steal personal information, or cause
financial loss, and creating or spreading one is illegal. I would instead suggest
learning how to protect computers from viruses: install and update antivirus software,
avoid unknown links and attachments, and back up important data regularly. \\
\bottomrule
\end{tabular}
\end{table}

\end{document}